\pdfoutput=1

\documentclass[11pt]{article}

\usepackage[final]{acl}

\usepackage{times}
\usepackage{latexsym}

\usepackage[T1]{fontenc}

\usepackage[utf8]{inputenc}

\usepackage{microtype}

\usepackage{inconsolata}

\usepackage{graphicx}
\usepackage{amsmath} 
\usepackage{mathtools}
\usepackage{amsmath}
\usepackage{amssymb}
\usepackage{amsfonts}
\usepackage{amstext}
\usepackage{amsthm}
\usepackage{booktabs}
\usepackage{enumitem}
\usepackage{url}
\usepackage{algorithm}
\usepackage{algpseudocode}

\title{BlackboxNLP-2025 MIB Shared Task: Improving Circuit Faithfulness via Better Edge Selection}

\author{
  Yaniv Nikankin\thanks{Equal contribution.} \quad
  Dana Arad\footnotemark[1] \quad
  Itay Itzhak\footnotemark[1] \\
  \textbf{Anja Reusch} \quad
  \textbf{Adi Simhi} \quad 
  \textbf{Gal Kesten-Pomeranz} \quad
  \textbf{Yonatan Belinkov} 
  \\
  Technion – Israel Institute of Technology \\[1ex]
  \{\href{mailto:yaniv.n@campus.technion.ac.il}{\texttt{yaniv.n}},
  \href{mailto:danaarad@campus.technion.ac.il}{\texttt{danaarad}},
  \href{mailto:itay.itzhak@campus.technion.ac.il}{\texttt{itay.itzhak}}\}\texttt{@campus.technion.ac.il} \\
}

\begin{document}
\maketitle
\begin{abstract}
One of the main challenges in mechanistic interpretability is circuit discovery---determining which parts of a model perform a given task.
We build on the Mechanistic Interpretability Benchmark (MIB) and propose three key improvements to circuit discovery. 
First, we use bootstrapping to identify edges with consistent attribution scores. 
Second, we introduce a simple ratio-based selection strategy to prioritize strong positive-scoring edges, balancing performance and faithfulness.
Third, we replace the standard greedy selection with an integer linear programming formulation.
Our methods yield more faithful circuits and outperform prior approaches across multiple MIB tasks and models. Our code is available at: \url{https://github.com/technion-cs-nlp/MIB-Shared-Task}.

\end{abstract}

\section{Introduction}
Mechanistic interpretability has gained recent popularity due to its progress in characterizing the internal mechanisms of AI models \cite{saphra2024mechanistic, Rai2024APR}. 
A popular paradigm aims to uncover \emph{circuits}, subgraphs of the model's computation graph that are responsible for specific tasks \cite{olah2020zoom}. 
However, discovering optimal circuits that are as small as possible while matching the original model's behavior remains an open challenge.
To this end, the recent Mechanistic Interpretability Benchmark (MIB) \cite{mueller2025mib} was proposed to offer a standardized framework for evaluating circuit discovery methods.

A typical circuit discovery pipeline consists of two stages: (1) obtaining scores for the full set of graph components (nodes, edges, etc.), and (2) selecting a subset of the components that constitute the circuit. 
Prior work has largely focused on developing improved scoring methods for the first stage, such as Edge Attribution Patching (EAP) \citep{nanda2023attribution} and EAP with Integrated Gradients (EAP-IG) \citep{hanna2024have}, while relying on a greedy selection algorithm for the second stage.
Using this setup, \citet{mueller2025mib} demonstrated that EAP-IG scores led to the top-preforming subgraphs.

In this work, as a part of the BlackboxNLP 2025 Shared Task \cite{mib2025blackboxnlp}, we focus on the second stage of circuit discovery and suggest several improvements for building a circuit given EAP-IG scores. We rely on a few key observations. 
First, EAP-IG scores can vary across data samples from the same task, with some edges receiving both negative and positive values in different samples.
The score sign represents a significant property: positive-scoring components contribute positively to the model’s performance on the task, while negative scores indicate a negative impact, such as the negative name mover heads from~\citet{wang2023interpretability}.
By bootstrapping the scores across resamples of the training data, we are able to identify edges with consistent score signs and filter out unstable ones. 

Second, we find that selecting edges by score magnitude alone, ignoring sign, often yields circuits that misrepresent the model’s original behavior.
To address this, we introduce a simple ratio-based strategy: select a fixed proportion of top-positive edges, and the rest by absolute value. 
This approach allows finer control over the balance of edge types and improves circuit faithfulness. 

Lastly, we formulate circuit construction as an Integer Linear Programming (ILP) optimization problem, instead of using the naive greedy solution. 

Overall, we show that different combinations of our methods, tailored to the different faithfulness objectives proposed by MIB, yield improved performance over the leading approaches.

\begin{table*}[t]
    \centering
    \begin{tabular}{@{}l@{} @{\hspace{7pt}}c @{\hspace{7pt}}r @{\hspace{5pt}}r @{\hspace{5pt}}r @{\hspace{5pt}}r r r @{\hspace{5pt}}r @{\hspace{5pt}}r r @{\hspace{5pt}}r r}
    \toprule
    & \multicolumn{3}{c}{IOI} & \multicolumn{2}{c}{MCQA} & \multicolumn{1}{c}{ARC (E)} \\\cmidrule(lr){2-4}\cmidrule(lr){5-6}\cmidrule(lr){7-7}
    \textbf{Method} & GPT-2 & Qwen-2.5 & Gemma-2 & Qwen-2.5 & Gemma-2 & Gemma-2 \\
    \midrule
    \textsf{CMD} Baseline (Greedy) & 0.0308 & 0.0374 & \textbf{0.0658} & 0.1846 & \textbf{0.0880} & 0.0458 \\
    \textsf{CMD} (ILP + PNR) & \textbf{0.0294} & \textbf{0.0370} & 0.0760 & \textbf{0.1820} & 0.0907 & \textbf{0.0451} \\   
    \midrule
    \textsf{CPR} Baseline (Greedy) & 2.4901 & 2.2658 & \textbf{5.6155} & 1.0612 & 1.8769 & \textbf{1.9104} \\
    \textsf{CPR} (ILP + Bootstrapping) & \textbf{2.5061} & \textbf{2.5092} & 5.4516 & \textbf{1.0926} & \textbf{1.9145} & 1.8918 \\
    \bottomrule
    \end{tabular}
    \caption{Comparison of our chosen methods against the baseline for \textsf{CMD} (lower is better) and \textsf{CPR} (higher is better) metrics on the public test sets.}
    \label{tab:main_scores}
\end{table*}

\section{Motivation}




Circuit discovery follows a two-stage pipeline: scoring and selection. In the scoring stage, each component of the model’s computation graph is given an importance score; in this submission we consider edges as our basic computation components and utilize Edge Attribution Patching with Integrated Gradients (EAP-IG) as our base scoring metric.
We use EAP-IG as it obtained the best score on MIB across most of the models and tasks. 

While EAP-IG is typically computed once over the entire training set, applying it to multiple data subsets reveals an interesting pattern.
In GPT-2$+$IOI and Qwen-2.5$+$MCQA, we find that $9.1$\% and $6.5$\% of edges, respectively, exhibit sign instability with both positive and negative scores across 10 samples.\footnote{Only edges with non-zero mean scores $|\mu| > 10^{-6}$.}
Since sign indicates whether an edge contributes positively or negatively to circuit performance, instability may signal noisy attribution scores. 
By applying \textbf{bootstrapping}, we can filter out unstable edges before constructing the graph. 

After scoring each edge, the selection stage follows in which a subset of edges is selected to form the circuit. 
In this stage, \citeauthor{mueller2025mib}\ employ the approach of using a greedy, layer-by-layer algorithm~\citep{hanna2024have}.
The greedy algorithm starts from the output layer and works backward, adding the top-ranked edges needed to compute the activations of already selected components.
The resulting graph is pruned to remove non-connected components, resulting in a connected subgraph.

To evaluate circuits, \citeauthor{mueller2025mib}\ define two metrics based on the faithfulness curve with respect to circuit size. 
The integrated circuit performance ratio (\textsf{CPR}) measures how well the method identifies components that positively contribute to the model’s performance on a task, and is defined as the area under the curve. 
In contrast, the integrated circuit-model distance (\textsf{CMD}) quantifies how closely the circuit approximates the model’s overall behavior, including positive and negative contributions, and is defined as the area between the faithfulness curve and the optimal value of $1$. 


Thus, in the MIB implementation edges are ranked differently between the two metrics. 
For \textsf{CPR} edges are simply ranked by their scores, whereas for \textsf{CMD} the ranking is based on their absolute scores. 
We observe that for \textsf{CMD}, this may lead to over-selection of negatively scoring edges, which can degrade the circuit's faithfulness.
To mitigate this, we propose a \textbf{positive-negative ratio (PNR)} strategy: first select a given percentage of top-positive edges, then apply the usual selection.

While the greedy algorithm is fast and guarantees a valid circuit, it relies solely on local decisions and may result in suboptimal edge selection.
Thus, we formulate circuit selection as an \textbf{integer linear program (ILP)} for globally optimal subset selection under structural and budget constraints. 


\section{Method}\label{sec:method}
In this section, we describe three methods for improving circuit construction, which can be applied individually or in combination. 

\subsection{Bootstrapped Confidence Filtering}

We use bootstrapping to identify edges with consistently-signed attribution scores, since these are more likely to reflect meaningful structure in the model.
For a dataset of size $N$, we sample with replacement to obtain $\tau$ sets with $N$ samples each. 
For each edge $e$, we collect a set of scores $\{a_1, \ldots, a_\tau\}$ from the $\tau$ bootstrap runs. We compute the sample mean $\mu_e$ and standard deviation $\sigma_e$, then construct a two-sided confidence interval:

\begin{equation}\label{eq:confidence_interval}
    \mu_{e} \pm z \cdot \frac{\sigma_e}{\sqrt{\tau}},
\end{equation}
where $\mu_e$ is the mean score across the bootstrap samples, $\sigma_e$ is the sample standard deviation, and $z$ is the standard normal quantile corresponding to the desired confidence level ($z = 1.96$ for 95\%).

An edge is retained if its confidence interval in Equation~\ref{eq:confidence_interval} lies entirely above or below a fixed significance threshold, depending on $\mu_e$'s sign, with $\mu_{e}$ serving as the final edge score.

\subsection{Positive-Negative Ratio (PNR)}

We select edges in two phases defined by the PNR value, a real number in the range $[0, 1]$. 
Given $k$ as the maximum number of edges in the circuit:
\begin{enumerate}[nosep]
    \item Select $\lceil \text{PNR} \cdot k \rceil$ edges from the top positively scored edges, sorted by raw signed-score.
    \item Select the remaining $k - \lceil \text{PNR} \cdot k \rceil$ edges from the top remaining edges, sorted by absolute score (positive or negative).
\end{enumerate}

Hence, $\text{PNR}$ sets the minimum fraction of positively contributing edges in the circuit.

\subsection{Integer Linear Programming (ILP)}
Lastly, we replace the greedy algorithm by formulating graph construction as an ILP problem \cite{wolsey1998integer}. 

Formally, we define the computation graph as a multi-edge directional graph, $G = (V, E)$, with a scoring function, $a: E \rightarrow \mathbb{R}$.
$G$ has a unique source node $y_s$ with no incoming edges ($d_{in}(y_s) = 0$), and a unique target node $y_t$ with no outgoing edges ($d_{out}(y_t) = 0$).
Let $x_e \in \{0,1\}$ indicate whether edge $e = (u,v,w) \in E$ is selected, and $y_v \in \{0,1\}$ indicate whether node $v \in V$ is used.

Given a budget $k$ on the maximum number of edges in the circuit, the ILP maximizes the total score of the selected edges (Equation~\ref{eq:objective}) under the following constraints:
\begin{subequations}\label{eq:ilp}
\begin{gather}
\max \quad \sum_{e \in E} a(e) \cdot x_e \label{eq:objective}
\end{gather}
\vspace{-1em}
\begin{align}
\text{s.t.} \quad & \sum_{\phantom{111|}e \in E\phantom{111}} x_e \leq k \label{eq:budget} \\
& y_s = y_t = 1 \label{eq:st} \\
& x_{(u,v,w)} \leq \min\{y_u, y_v\}, \text{\footnotesize $\forall (u,v,w) \in E$} \label{eq:en-consistency} \\
& \sum_{\substack{\phantom{1}(u,v,w) \in E}} x_{(u,v,w)} \geq y_u,  \text{\footnotesize $\forall u \in V$} \label{eq:u-connectivity} \\
& \sum_{\substack{\phantom{1}(u,v,w) \in E}} x_{(u,v,w)} \geq y_v, \text{\footnotesize $\forall v \in V$} \label{eq:v-connectivity} \\
& \sum_{\substack{e \in E, a(e) > 0}} x_e \geq \operatorname{PNR} \cdot k\label{eq:pnr}
\end{align}
\end{subequations}
where the constraints correspond to:
\begin{itemize}[nosep]
    \item \textbf{Edge budget (\ref{eq:budget}):} select at most $k$ edges.
    \item \textbf{Source and target (\ref{eq:st}):} source node $y_{s}$ and target node $y_{t}$ must be selected. 
    \item \textbf{Node-edge consistency (\ref{eq:en-consistency}):} if $(u,v,w)$ is selected, both $u$ and $v$ must be selected.
    \item \textbf{Connectivity (\ref{eq:u-connectivity}, \ref{eq:v-connectivity}):} every used non-source node has at least one incoming selected edge; every used non-target node has at least one outgoing selected edge.
    \item \textbf{PNR (\ref{eq:pnr}):} (when using ILP with PNR) the number of positive-scoring edges should exceed the PNR value.
\end{itemize}

\begin{table*}[t]
    \centering
    \begin{tabular}{@{}l@{} @{\hspace{7pt}}c @{\hspace{7pt}}r @{\hspace{5pt}}r @{\hspace{5pt}}r @{\hspace{5pt}}r r r @{\hspace{5pt}}r @{\hspace{5pt}}r r @{\hspace{5pt}}r r}
    \toprule
    & \multicolumn{3}{c}{IOI} & \multicolumn{2}{c}{MCQA} & \multicolumn{1}{c}{ARC (E)} \\\cmidrule(lr){2-4}\cmidrule(lr){5-6}\cmidrule(lr){7-7}
    \textbf{Method} & GPT-2 & Qwen-2.5 & Gemma-2 & Qwen-2.5 & Gemma-2 & Gemma-2 \\
    \midrule
    Greedy & 0.0411 & 0.0254 & 0.0564 & 0.1403 & 0.1234 & \textbf{0.0417} \\
    \midrule
    ILP & \underline{0.0370} & \underline{0.0242} & 0.0646 & \textbf{\underline{0.1348}} & 0.1253 & 0.0477 \\
    Bootstrapping & 0.0965 & 0.2849 & \underline{0.0350} & 0.3260 & \underline{0.0906} & 0.0489 \\
    PNR & \textbf{\underline{0.0221}} & \textbf{\underline{0.0217}} & 0.0606 & 0.1484 & \underline{0.1058} & 0.0548 \\
    \midrule
    ILP + Bootstrapping & 0.0961 & 0.2844 & \textbf{\underline{0.0295}} & 0.3290 & \underline{0.0902} & 0.0438 \\
    ILP + PNR & \underline{0.0370} & \underline{0.0242} & 0.0590 & \textbf{\underline{0.1348}} & \underline{0.1047} & 0.0477 \\
    Bootstrapping + PNR & 0.0495 & 0.1572 & \underline{0.0336} & \underline{0.1404} & \textbf{\underline{0.0586}} & 0.0489 \\
    \midrule
    ILP + Bootstrapping + PNR & 0.0886 & 0.2734 & \textbf{\underline{0.0295}} & 0.2582 & \underline{0.0879} & 0.0427 \\
    \bottomrule
    \end{tabular}
    \caption{\textsf{CMD} scores across all combinations of our methods on the public validation sets (lower is better). We \textbf{bold} the best method per column and \underline{underline} any result better than the greedy baseline.}
    \label{tab:cmd-method-ablations}
    \vspace{-3pt}
\end{table*}

\begin{table*}[t]
    \centering
    \begin{tabular}{@{}l@{} @{\hspace{7pt}}c @{\hspace{7pt}}r @{\hspace{5pt}}r @{\hspace{5pt}}r @{\hspace{5pt}}r r r @{\hspace{5pt}}r @{\hspace{5pt}}r r @{\hspace{5pt}}r r}
    \toprule
    & \multicolumn{3}{c}{IOI} & \multicolumn{2}{c}{MCQA} & \multicolumn{1}{c}{ARC (E)} \\\cmidrule(lr){2-4}\cmidrule(lr){5-6}\cmidrule(lr){7-7}
    \textbf{Method} & GPT-2 & Qwen-2.5 & Gemma-2 & Qwen-2.5 & Gemma-2 & Gemma-2 \\
    \midrule
    Greedy & 2.9302 & 2.2553 & \underline{5.4410} & 1.2421 & 1.8217 & 1.9217 \\
    \midrule
    ILP & 2.9153 & 2.1928 & 5.3489 & \textbf{\underline{1.3844}} & \underline{1.8240} & 1.9197 \\
    Bootstrapping & \textbf{\underline{3.1865}} & \textbf{\underline{2.6196}} & \textbf{\underline{5.6091}} & \underline{1.2548} & \underline{1.9142} & \underline{1.9846} \\
    \midrule
    ILP + Bootstrapping\phantom{ + PNR}  & \underline{3.1772} & \underline{2.5894} & \underline{5.4410} & \underline{1.3510} & \textbf{\underline{1.9261}} & \textbf{\underline{2.0099}} \\
    \bottomrule
    \end{tabular}
    \caption{\textsf{CPR} scores across all combinations of our methods on the public validation sets (higher is better). We \textbf{bold} the best method per column and \underline{underline} any result better than the greedy baseline.}
    \label{tab:cpr-method-ablations}
    \vspace{-3pt}
\end{table*}

\section{Experimental Setup}\label{sec:exp_setup}

MIB provides a standardized framework for evaluating circuit discovery methods across four models and four tasks. 
In this submission, we focus on a subset of these models and tasks due to computational limitations of the ILP problems, and include results on Gemma-2 2B~\citep{team2024gemma}, Qwen-2.5 0.5B~\citep{yang2024qwen2}, and GPT-2 Small~\citep{radford2019language} on indirect object identification (IOI)~\citep{wang2022interpretability}, multiple-choice question answering (MCQA)~\citep{wiegreffe2024answer}, and the easy partition of the AI2 Reasoning Challenge (ARC-E)~\citep{clark2018think}. 




We used the validation sets to select the best combination of methods and hyper-parameters for each metric, and additionally report results of the leading method on the test sets. 
%
%
%
%
Additional implementation details are described in Appendix~\ref{app:implenetations}. 


\section{Results}\label{sec:results}
Our main results are displayed in Table~\ref{tab:main_scores}. We evaluate using both \textsf{CMD} (lower is better) and \textsf{CPR} (higher is better), reporting our best-performing combination of proposed methods against the greedy graph building baseline \cite{mueller2025mib}.
For \textsf{CMD} evaluation, we employ ILP to construct optimal graphs, combined with PNR selection, which prioritizes positive edges.
The PNR value varies per model-task combination and was chosen for each combination individually, to account for task-specific distributions. 
This approach enhances faithfulness towards the optimal threshold of $1.0$, particularly for smaller subgraphs.

For \textsf{CPR} evaluation, where negative edges are penalized due to the objective of maximizing faithfulness, we replace PNR with bootstrapping to consistently retain positive edges.
We use $\tau=10$ bootstrap iterations.
We choose these combinations of methods as they achieve better results than the baseline across the widest range of analyzed models and tasks (see Section~\ref{sec:ablations}).
Our proposed methods demonstrate improvements over the baseline across almost all models and tasks evaluated.

\subsection{Ablations}
\label{sec:ablations}
We conduct an ablation study to evaluate our method design choices. We ablate both the combination of methods and method-specific hyper-parameters.
Tables \ref{tab:cmd-method-ablations} and \ref{tab:cpr-method-ablations} present the \textsf{CMD} and \textsf{CPR} results, respectively, across different model and task combinations. 
All ablations are performed on the MIB validation sets.
Additional ablation studies on the number of bootstraps and PNR ratio values are provided in Appendix~\ref{app:further-ablations}.

\section{Discussion and Limitations}

Our work focused on the second stage of circuit discovery: using edge scores to select a fully connected sub-graph.
This aspect has received limited attention in prior work, with most approaches relying on naive top-$n$ selection or greedy algorithms \cite{hanna2024have, conmy2023towards}.
We demonstrated that the greedy algorithm can be improved through techniques such as ILP, bootstrapping, and PNR. 
However, our methods come with important limitations.
First, ILP optimization scales poorly with edge count, limiting applicability to larger models while providing only modest faithfulness gains.
Second, the optimal ratio of positive edges varies by models and tasks, requiring task-specific tuning.
These limitations increase the computational overhead required to achieve higher-faithfulness graphs from existing edge scores.

Despite these limitations, our results show that principled edge selection improves faithfulness and enables more robust circuit discovery.

While early experiments showed that ILP significantly outperformed greedy methods at maximizing total edge scores, this did not translate to significantly higher faithfulness scores. 
This suggests a gap between state-of-the-art attribution methods \citep{hanna2024have} and the ``ground truth'' edge importance scores. Better score attribution methods could potentially unlock the full benefits of using ILP as an optimal solution to graph building.

\section*{Acknowledgments}
This research was supported by an Azrieli Foundation Early Career Faculty Fellowship and Open Philanthropy.
D.A. is supported by the Ariane de Rothschild Women Doctoral Program.
A.R. was funded through the Azrieli international postdoctoral fellowship and the Ali Kaufman postdoctoral fellowship. 
A.S is supported by the Council for Higher Education (VATAT) Scholarship for PhD students in data science and artificial intelligence.
This research was funded by the European Union (ERC, Control-LM, 101165402). 
Views and opinions expressed are however those of the author(s) only and do not necessarily reflect those of the European Union or the European Research Council Executive Agency.
Neither the European Union nor the granting authority can be held responsible for them.

\bibliography{custom}

\newpage
\appendix

\begin{table*}[h]
    \centering
    \begin{tabular}{@{}l@{} @{\hspace{7pt}}c@{} @{\hspace{7pt}}c @{\hspace{7pt}}r @{\hspace{5pt}}r @{\hspace{5pt}}r @{\hspace{5pt}}r r r @{\hspace{5pt}}r @{\hspace{5pt}}r r @{\hspace{5pt}}r r}
    \toprule
    & & \multicolumn{3}{c}{IOI} & \multicolumn{2}{c}{MCQA} & \multicolumn{1}{c}{ARC (E)} \\\cmidrule(lr){3-5}\cmidrule(lr){6-7}\cmidrule(lr){8-8}
    \textbf{Method} & \textbf{n} & GPT-2 & Qwen-2.5 & Gemma-2 & Qwen-2.5 & Gemma-2 & Gemma-2 \\
    \midrule
    Bootstrapping & 5 & 3.1617 & 2.5794 & \textbf{5.6895} & \textbf{1.3918} & 1.8130 & 1.9615 \\
    Bootstrapping & 10 & 3.1865 & \textbf{2.6196} & 5.6091 & 1.2548 & \textbf{1.9142} & \textbf{1.9846} \\
    Bootstrapping & 15 & \textbf{3.1988} & 2.5527 & 5.4732 & 1.0808 & 1.9136 & 1.9616 \\
    \bottomrule
    \end{tabular}
    \caption{\textsf{CPR} scores (higher is better) across different amounts of bootstrap iterations $n$. \textbf{Bold} indicates best result per task-model combination.}
    \label{tab:bootstrap-n-ablations}
\end{table*}

\begin{table*}[h]
    \centering
    \begin{tabular}{@{}l@{} @{\hspace{7pt}}c @{\hspace{7pt}}r @{\hspace{5pt}}r @{\hspace{5pt}}r @{\hspace{5pt}}r r r @{\hspace{5pt}}r @{\hspace{5pt}}r r @{\hspace{5pt}}r r}
    \toprule
    & \multicolumn{3}{c}{IOI} & \multicolumn{2}{c}{MCQA} & \multicolumn{1}{c}{ARC (E)} \\\cmidrule(lr){2-4}\cmidrule(lr){5-6}\cmidrule(lr){7-7}
    \textbf{Method} & GPT-2 & Qwen-2.5 & Gemma-2 & Qwen-2.5 & Gemma-2 & Gemma-2 \\
    \midrule
    ILP + PNR $0.3$  & 0.0416  & 0.0322 & \textbf{0.0590} & 0.1852 & \textbf{0.1047} & \textbf{0.0477} \\
    ILP + PNR $0.4$  & 0.0416  & 0.0322 & \textbf{0.0590} & 0.1852 & \textbf{0.1047} & \textbf{0.0477} \\
    ILP + PNR $0.45$  & 0.0416 & 0.0322 & \textbf{0.0590} & 0.1852 & \textbf{0.1047} & \textbf{0.0477} \\
    ILP + PNR $0.5$  & 0.0416  & 0.0322 & \textbf{0.0590} & 0.1852 & 0.1068 & \textbf{0.0477} \\
    ILP + PNR $0.55$  & 0.0416 & 0.0322 & \textbf{0.0590} & 0.1852 & 0.1068 & \textbf{0.0477} \\
    ILP + PNR $0.6$  & \textbf{0.0370}  &\textbf{ 0.0242} & 0.0646 & \textbf{0.1348} & 0.1253 & \textbf{0.0477} \\
    ILP + PNR $0.7$  & \textbf{0.0370}  & \textbf{0.0242} & 0.0646 & \textbf{0.1348} & 0.1253 & \textbf{0.0477} \\
    ILP + PNR $0.8$  & \textbf{0.0370}  & \textbf{0.0242} & 0.0646 & \textbf{0.1348} & 0.1253 & \textbf{0.0477} \\
    ILP + PNR $0.9$  & \textbf{0.0370}  & \textbf{0.0242} & 0.0646 & \textbf{0.1348} & 0.1253 & \textbf{0.0477} \\
    \bottomrule
    \end{tabular}
    \caption{\textsf{CMD} scores (lower is better) across different PNR values. \textbf{Bold} indicates best result per task-model combination.}
   \label{tab:pnr-ablations}
\end{table*}

\section{Hyper-parameter Ablations}
\label{app:further-ablations}

We report additional ablations on method-specific hyperparameters. 
For bootstrapping, we tested different numbers of iterations with $\tau \in \{5, 10, 15\}$. For the PNR method, we explored values of $\text{PNR} \in \{0.3, 0.4, 0.45, 0.5, 0.55, 0.6, 0.7, 0.8, 0.9\}$.
Tables \ref{tab:bootstrap-n-ablations} and \ref{tab:pnr-ablations} present the results for bootstrapping and PNR ablations, respectively. These results motivated our selection of $n=10$ bootstrapping iterations and model-task-specific PNR values that yielded the best validation performance.

\section{Implementation Details}
\label{app:implenetations}
Our implementation is based on the code provided in MIB.\footnote{\url{https://github.com/hannamw/MIB-circuit-track}}
The EAP-IG attribution scores were computed using the implementation by \citeauthor{hanna2024have}. 
Our ILP approach employs Pulp~\cite{mitchell2011pulp} as a linear integer programming modeler, using Cbc~\cite{cbc} as the solver.

\end{document}